\newcommand{\avmplus}[1]{{\setlength{\arraycolsep}{0.8mm}	
                       \renewcommand{\arraystretch}{0.7}
                       \left[ 			
                       \begin{array}{l}
                       \\[-2mm] #1 \\[-2mm] \\
                       \end{array} 		
                       \right]
                    }}
\newcommand{\att}[1]{{\mbox{\scriptsize {\bf #1}}}}
\newcommand{\attval}[2]{{\mbox{\scriptsize {\sc #1}}\ =\ {{#2}}}}
\newcommand{\attvaltyp}[2]{{\mbox{\scriptsize{\sc #1}}\ =\ {\myvaluebold{#2}}}}
\newcommand{\attvalshrunktyp}[2]{{\mbox{\scriptsize{\sc #1}}\ =\ {\boxvaluebold{#2}}}}
\newcommand{\myvaluebold}[1]{{\mbox{\scriptsize {\bf #1}}}}
\newcommand{\boxvaluebold}[1]{{\fbox{\scriptsize {\bf #1}}}}

\documentstyle[12pt,a4,colon]{article}

\newenvironment{exquote}{\begin{quote}\footnotesize}{\end{quote}}

\begin{document}

\begin{center}
{\Large\bf Bi-Lexical Rules for Multi-Lexeme Translation in Lexicalist MT}\\[1em]
{\bf Arturo Trujillo} \\
SCMS, The Robert Gordon University\\ St Andrew Street, Aberdeen AB1 1HG\\ 
Scotland \\ {\tt iat@scms.rgu.ac.uk}\\cmp-lg/9508006\\
{\small (in {\em Proceedings of the Sixth International Conference on 
Theoretical and Methodological Issues in Machine Translation -- TMI-95}, 
Leuven, Belgium, July 1995, pp. 48--66)}\\[1em]
\end{center}

\abstract{
The paper presents a prototype lexicalist Machine Translation system (based on
the so-called `Shake-and-Bake' approach of \cite{whitelock92}) consisting of an
analysis component, a dynamic bilingual lexicon, and a generation component,
and shows how it is applied to a range of MT problems. Multi-Lexeme
translations are handled through bi-lexical rules which map bilingual lexical
signs into new bilingual lexical signs. It is argued that much translation can
be handled by equating translationally equivalent lists of lexical signs,
either directly in the bilingual lexicon, or by deriving them through
bi-lexical rules.  Lexical semantic information organized as Qualia structures
\acite{pustejovsky91} is used as a mechanism for restricting the domain of
the rules.
}

\normalsize
\section{Introduction}

Transfer based approaches to machine translation (MT) involve three main
phases: analysis, transfer and generation. During analysis, the syntactic and
semantic structure of a sentence is made explicit through a source language
(SL) grammar and semantic processing modules. The result of analysis is one or
more syntactic and semantic representations which are used to construct a
syntactic and/or semantic representation in the target language (TL) through a
series of transfer rules and a bilingual lexicon. From this representation a TL
sentence is generated based on some form of mapping procedure, usually
exploiting the TL grammar \footnote{
While this definition of transfer systems is current in most MT discussions, it
has been challenged \acite{kayetal94} on the basis that the interlingua-transfer
distinction, that is, the distinction between systems which construct language
independent representations and systems which do not, is artificial and that in
fact the two paradigms simply represent different aspects of the same problem.
While we agree with this observation, many systems at present start with
an interlingua or a transfer architecture and then incorporate solutions from
the alternative paradigm. We therefore maintain the distinction, at least for
the purposes of this paper.}.

In this paper we describe a prototype implementation of a transfer MT system
based on the lexicalist MT (LMT) approach of \cite{whitelock92}, also
known as `Shake-and-Bake' (SB).  For our implementation we have extended the
original SB formulation by postulating bilingual lexical rules (bi-lexical rules
henceforth)
which dynamically expand the bilingual lexicon in order to
extend its functionality.  This allows us to uniformly treat mono- and
multi-lexeme translations in a variety of contexts.

We describe the main characteristics of the LMT approach. This is followed by a
description of the problems posed by certain multi-lexeme translations, and of
how bi-lexical rules, in conjunction with lexical semantic information provide
a framework for overcoming these problems. We then point out some limitations
in our approach and give some idea as to the status of our implementation.
 
\section{Lexicalist Machine Translation}

In its original formulation, LMT consists of three main phrases: analysis,
lexical-semantic transfer and generation.  The analysis phase involves parsing
the input sentence to produce an output bag or multiset of SL lexical signs
instantiated with sufficient information to permit appropriate translation.
Transfer maps these signs into a TL bag through the bilingual lexicon in which
sets of source and target lexical signs are placed in translation
correspondence. Generation consists of finding an ordering of the TL bag
which satisfies the constraints imposed by the TL grammar.  Normally,
generation involves a modified parser which ignores ordering information
\acite{brew92,popowich95} although other approaches are also possible
\acite{poznanskietal95}.

\subsection{Notation}

We introduce some notation through a simple example of our implementation.
Since we will not be concerned with quantification nor scoping, we adopt a
simplified transfer representation. If quantification and scope were to be
included, however, a mechanism along the lines of \cite{franketal95} and
\cite{copestakeetal95b} may be followed in order to preserve the recursiveless
nature of lexicalist transfer.

Our lexical signs broadly follow the signs of \cite{pollardetal87} although our
work seems adaptable to the signs of \cite{pollardetal94}. The implementation
is based on the Typed Features Structures (TFSs) of the Acquilex LKB
\acite{copestakeetal93} from where we borrow our notation. Consider the (simplified)
lexical entry for `John':
\begin{quote}
{\scriptsize
$\avmplus{\att{proper-name}\\
\attvaltyp{orth}{John}\\
\attval{syn}{\avmplus{\att{syn}\\
             \attvaltyp{agr}{3sg}}}\\
\attvalshrunktyp{qualia}{qualia}\\
\attvaltyp{sem}{john1(x)}}$}
\end{quote}
In this TFS, features are written in small capitals, while types are in bold
face. To make TFSs easier to read, detail may be hidden by `shrinking' a TFS;
this is indicated with a box around the type of the TFS (e.g.
$\boxvaluebold{qualia}$ above). TFSs of type {\bf qualia} encode lexical
semantic information based on the Qualia structures of \cite{pustejovsky91}.
For the semantic representation of proper names we assume a predicate treatment
following the arguments of \cite[225]{devlin91}. A bilexical entry for `John
-- {\em Juan}' would be:
\begin{quote}
{\scriptsize
$\avmplus{\att{proper-name}\\
\attvaltyp{orth}{John}\\
\attval{syn}{\avmplus{\att{syn}\\
             \attvaltyp{agr}{3sg}}}\\
\attvalshrunktyp{qualia}{qualia}\\
\attvaltyp{lang}{english}\\
\attvaltyp{sem}{john1(x)}} 
\leftrightarrow
\avmplus{\att{proper-name}\\
\attvaltyp{orth}{Juan}\\
\attval{syn}{\avmplus{\att{syn}\\
             \attvaltyp{agr}{3sg}}}\\
\attvalshrunktyp{qualia}{qualia}\\
\attvaltyp{lang}{spanish}\\
\attvaltyp{sem}{juan1(x)}}$}
\end{quote}

For reasons of space and convenience, we will abbreviate the above lexical sign
and bilexical entry to
\begin{exquote}
john1$_{x}$\\
john1$_{x}$ $\leftrightarrow$ juan1$_{x}$\\
\end{exquote}
respectively, where the subscripts correspond to the argument variable. It
should be emphasised, however, that this abbreviated notation implicitly
includes syntactic and semantic information which may be accessed during
transfer or generation.

To exemplify LMT, consider the translation of `John likes Mary'. Analysis
results in a list\footnote{We use lists of SL lexical items, instead of bags as
is done in SB, to avoid certain inefficiencies caused by the nature of
lexicalist transfer \acitec[221]{gareyetal79}.} of lexical signs the semantics
of which will contain shared variables:
\begin{exquote}
john1$_{x}$ love1$_{e,x,y}$ mary1$_{x}$
\end{exquote}
The (tenseless) FOL formula corresponding to this expression is $\exists exy$.
john1($x$) \& love1($e,x,y$) \& mary1($y$), but since quantification and scope
will be ignored they will be omitted from our examples; furthermore,
coordination will be assumed between predicates unless otherwise stated.

Before transfer, a process similar to skolemization is applied to the transfer
representation in order to replace variables by constants.  The purpose of this
operation is to prevent spurious bindings during lexicalist generation, as will
become clearer later. The result of analysis is a list of lexical signs with
translationally relevant relationships expressed by shared constants (indicated
by integers in our  notation):
\begin{exquote}
john1$_{1}$ love1$_{2,1,3}$ mary1$_{3}$
\end{exquote}
The transfer step uses the source side of the bilexicon (possibly expanded by
bilingual lexical rules as described below) to derive a total cover of the SL
list \acitec[221]{gareyetal79} (a total cover is a division of a set into a
number of allowed subsets such that every element in the set is a member of
exactly one subset; we extend the term here to apply it to lists). The
bilexicon below enables construction of an appropriate TL bag:
\begin{quote}
\footnotesize
john1$_{x}$ $\leftrightarrow$ juan1$_{x}$\\
mary1$_{x}$ $\leftrightarrow$ mar\'\i a1$_{x}$\\
love1$_{x,y,z}$ $\leftrightarrow$ amar1$_{x,y,z}$ a1$_{z}$
\end{quote}
(Tense is omitted in this example; a simplistic model has been adopted in which an
interlingua tense feature is passed from source to target verbs in the
bilexicon.)  Note that we include function words such as the Spanish case
marker {\em a} in the bilingual lexicon (and therefore in the transfer
representation). These words are treated as vacuous predicates
\acite{calderetal89} over the variable of the semantic head on which they
depend. For the present example, transfer results in the following TL bag:
\begin{quote}
\footnotesize
\{juan1$_{1}$ , amar1$_{2,1,3}$ , a$_{3}$ , mar\'\i a1$_{3}$\}
\end{quote}
Lexicalist generation involves reordering the TL bag to construct a valid TL
sentence. Since normally all permutations of the TL bag are attempted, the fact
that variables are replaced by constants ensures that arguments not shared
between predicates in the SL representation are not shared in the TL
representation either. This prevents {\em Mar\'\i a} from being the subject of
the sentence. The result of generation, after morphological synthesis, is:
\begin{quote}
\footnotesize
{\em Juan ama a Mar\'\i a}
\end{quote}

\subsection{Other Properties of LMT}

LMT encourages two useful properties: modularity and reversibility.  From an
engineering point of view, modularity is desirable because it can reduce
development and maintenance costs. By using sets of lexical signs as their
transfer representation, LMT systems can reduce the difficulties posed by
structural mismatches between two languages, thus increasing the independence
between source and target transfer representations. For example, transfer
systems adopting a recursive representation for transfer \acite{kaplanetal89},
as opposed to a non-recursive one \acite{copestakeetal95b}, may need additional
mechanisms for handling head switching \acite{kaplanetal93}. By contrast,
under a lexicalist approach, head switching can be handled purely
compositionally with minimal assumptions \acite{whitelock92}. 

Reversibility is an important property in bi-directional systems as it reduces
development costs. In LMT, grammars are fully reversible since they are used in
similar ways for analysis and generation: the difference is that during
lexicalist generation, ordering information is disregarded.  However, the
process is complete because the generator is guaranteed to generate
all the strings accepted by the TL grammar which satisfy the constraints
imposed by the TL bag.  Lexicalist generation is also sound because only
strings which satisfy the constraints of the TL grammar are constructed.  In
addition, termination is guaranteed if it is guaranteed for parsing since one
can at worst construct a generation algorithm which simply attempts all
permutations of the TL bag and then parses them in order to test whether they
are appropriate TL sentences.

\section{Multi-Lexical Translations}
\label{mul-tra-sec}

One of the reasons for transfer modules being expensive to construct is the
presence of complex transfer relations \acite{arnoldetal92,hutchinsetal92}. One
type of phenomena that leads to complex transfer in a number of systems may be
called multi-lexical translation. These are translations in which a phrase
cannot easily be translated through the translation of its parts. The
translation of idioms is an extreme case of this. For example, `kick the
bucket' translates as {\em estirar la pata} (Lit. `to stretch a leg') in
Spanish, even though there is no simple correspondence between the components
of each phrase (all translations in this paper are between English and Spanish
unless otherwise stated). For such constructions, structures corresponding to
the source and target phrases need to be equated either in the transfer module
\acite{schenk86} or in separate dictionaries \acite{sadleretal90} in many
systems. Other phenomena which may be loosely labelled multi-lexeme
translations include: lexical gaps such as `piece of advice' -- {\em consejo}
\acite{soleretal93}; support verb and category differences such as `to be
thirsty' -- {\em tener sed} (to have thirst) \acite{danlosetal92};
lexicalization patterns like `swim across the river Dee' -- {\em cruzar el
r\'\i o Dee nadando} \acite{talmy85}; conflational divergences as in `to stab
someone' -- {\em darle pu\~{n}aladas a alguien} \acite{dorr92}.

Phenomena such as idioms, lexical gaps and conflational divergences can be
tackled in LMT by equating sets of source and target lexical signs:
\begin{exquote}
a) kick1$_{e,s,o}$, the1$_{o}$, bucket1$_{o}$ $\leftrightarrow$ estirar1$_{e,s,o}$, la1$_{o}$, pata1$_{o}$\\
b) piece1$_{x}$, of1$_{x,y}$, advice1$_{y}$ $\leftrightarrow$ consejo1$_{x}$\\
c) stab1$_{e,s,o}$ $\leftrightarrow$ dar1$_{e,s,p,o}$ le1$_{o}$ pu\~nalada1$_{p}$ a1$_{o}$
\end{exquote}
(We include lexical signs for determiners, clitics and accusative markers as
predicates over the variable of their syntactic head; however, reasoning
formalisms may dispense with them.)  Note that we choose the variable of
`piece' on the English side as the argument variable on the Spanish side; if
phrases such as `a piece of good advice' are allowed, the Spanish side would be
{\em consejo}1$_{x\sqcup y}$, whose semantic argument would be unifiable with
both $x$ and $y$ to permit modifiers and heads to combine appropriately during
generation. 

To translate `John kicked the bucket', the SL transfer representation:
\begin{exquote}
john1$_{1}$ kick1$_{2,1,3}$ the1$_{3}$ bucket1$_{3}$ 
\end{exquote}
is covered by the bilexicon. The result is the union of the target side of all
the bilexical entries used in this process:
\begin{exquote}
\{juan1$_{1}$\} $\cup$ \{estirar1$_{2,1,3}$, la1$_{3}$, pata1$_{3}$\}
\end{exquote}
(We ignore the literal translation of the idiom.)  Generation then proceeds via
the Spanish grammar and bag generator.

In the case of the other multi-lexeme translations mentioned the difficulties
posed by varying lexical elements in part or all of the translation relation
cannot be easily handled in the original SB formulation. Consider for example
the case of `John is thirsty'; its Spanish translation, {\em Juan tiene sed}
(lit.  `John has thirst') differs from it in two main ways: the English
adjective translates into a Spanish noun, while the verb is not intuitively
felt to be the translation of {\em tener}. The problem for LMT based on
one-to-one transfer is that a literal translation into Spanish is incorrect
(*{\em Juan est\'a sediento}), and that even if TL filtering
\acite{alshawietal92} were used to eliminate such a sentence, the efficiency of
the system would be compromised and translation of unseen sentences would be
more error prone. Alternatively, an idiom-based translation in which the
bilexicon relates `be thirsty' and {\em tener sed} ignores important systematic
differences between the two languages:
\begin{exquote}
\begin{tabular}{ll}
John {\bf is thirsty} & Juan {\bf tiene sed}\\
John {\bf is hungry} & Juan {\bf tiene hambre}\\
John {\bf is lucky} & Juan {\bf tiene suerte}\\
John {\bf is angry} & Juan {\bf tiene rabia}\\
John {\bf is hot} & Juan {\bf tiene calor}\\
John {\bf is cold} & Juan {\bf tiene fr\'\i o}
\end{tabular}
\end{exquote}

We therefore argue that a one-to-one translation for such phrases is not
adequate but instead consider the highlighted phrases above as the correct
equivalences between the two languages. The task then, is to find a mechanism
for efficiently capturing regularities of this sort in the present framework.
There are a number of alternatives for achieving this.  We will consider three.

\subsection{Lexical Neutralization}
\label{lex-neu-sec}

The first possibility for handling multi-lexeme regularities in LMT is to
eliminate support verbs from the SL transfer representation altogether, and to
reintroduce them during generation. In this case, a semantic representation for
the sentences must be proposed.  For the sake of argument assume an
adjective-like intersective semantics for both the Spanish nouns {\em Juan} and
{\em sed} and the corresponding English noun and adjective: 
\begin{quote}
\footnotesize
SL: john1$_{1}$ thirsty1$_{1}$\\
TL: juan1$_{1}$ sed1$_{1}$
\end{quote}
Then, the bilexicon would include, among other things:
\begin{exquote}
thirsty1$_{x}$ $\leftrightarrow$ sed1$_{x}$\\
hungry1$_{x}$ $\leftrightarrow$ hambre1$_{x}$\\
{\em etc.}
\end{exquote}
Lexicalist transfer would apply these equivalences to construct an appropriate
TL bag. During Spanish bag generation, the appropriate support verb (i.e. {\em
tener}) would be introduced by inspection of monolingual lexical information
associated with {\em sed} \acite{danlosetal92}, from which correct
instantiation of the orthography of the TL sentence would ensue. A variation of
this strategy would be to use a partially instantiated lexical sign
corresponding to the English support verb:
\begin{quote}
\footnotesize
\{ john1$_{1}$ , support-verb$_{2,1,3}$ , thirsty$_{3}$ \}
\end{quote}
During transfer, the support verb is translated as a partially instantiated
support verb in Spanish. The generation algorithm would then be applied such that
monolingual constraints in the Spanish grammar fully instantiated the semantics
and orthography of this verb according to the support verb requirements of its
complement noun.

\subsection{Lexical Variables}
\label{lex-var-sec}

The second mechanism for capturing multi-lexeme regularities assumes
translation variables similar to those used in several transfer systems
\acite{alshawietal92,bechetal91,russelletal91}. If one represents transfer
variables by {\tt tr({\em $<$restrictions$>$})}, then the necessary bilexical
entry would be:
\begin{exquote}
be1$_{x,y,z}$, {\tt tr(}Adj$_{z}${\tt )} $\leftrightarrow$ 
tener1$_{x,y,z}$, {\tt tr(}Noun$_{z}${\tt )}
\end{exquote}
This entry states that `be' translates as {\em tener} as long as its complement
adjective translates as the complement noun of {\em tener}.  The transfer
algorithm is modified to accommodate the transfer variable by, for example,
recursively calling itself on the value of {\tt tr(}Adj$_{z}${\tt )}.
Generation, however, proceeds as before. A variation of this mechanism is to
use contextual rather than transfer variables. In this case, a particular
lexical context is specified which constraints translation equivalence in a
manner analogous to the way left and right contexts are used in morphological
rewriting rules \acite{kaplanetal94}. Thus, the transfer relation
\begin{exquote}
be1$_{x,y,z}$, (Adj$_{z}$) $\leftrightarrow$ 
tener1$_{x,y,z}$, (Noun$_{z}$)
\end{exquote}
would indicate that in the context of an adjective complement, `be' may
translate as {\em tener} or vice versa. The main difference between this and
the transfer variable variant is that the contextual elements, Adj and Noun,
can serve as context to multiple transfer relations within the same cover,
whereas this would not be possible with transfer variables. We will appeal to
contextual variables in Section \ref{bil-com-sec}.

The third mechanism uses bilingual lexical rules to map bilexical entries into
new bilexical entries. We have adopted this mechanism for certain multi-lexeme
translations because it allows the exploitation of monolingual lexical rules in
a motivated manner which integrates naturally with the LMT architecture, and
because it provides a framework in which to study differences between lexical
processes in different languages.

\section{Lexical and Bi-Lexical Rules}

The lexicon has taken a prominent place in several linguistic theories
\acite{pollardetal94,oehrleetal88}, not least because, given appropriate tools,
both general and idiosyncratic properties of language can be captured within a
uniform framework. Among the tools normally employed one finds lexical rules
\acite{dowty78,flickinger87,pollardetal94} and inheritance mechanisms
\acite{briscoeetal93a,flickingeretal92}. Lexical rules may be thought of as
establishing a relationship between lexical items such that given the presence
of one lexical item in the lexicon the existence of a further item may be
inferred. The regularities captured by lexical rules might include changes in
the subcategorization and control properties of a verb, the denotation of a
noun or the interpretation of a preposition. With the advent of lexically
oriented approaches to translation, it is worth considering whether and how the
generalizations captured by lexical rules might be exploited in MT.

In order to investigate this issue we have adopted the notion of a bi-lexical
rule. A bi-lexical rule \acite{trujillo92b,copestakeetal93} takes a bilexical
entry as input, and outputs a new bilexical entry. These rules may be seen as
expanding the bilexicon in order to increase its coverage; under this view,
they are somewhat analogous to lexical rules in that they reduce the number of
bilexical entries that need to be explicitly listed. Bi-lexical rules also
serve to capture lexical, syntactic and semantic regularities in the
translation between two languages by relating equivalent lexical processes
cross-linguistically.

\subsection{Simple Bi-lexical Rule}

We give a simple example of a bi-lexical rule before addressing the
multi-lexeme translations introduced earlier. Consider the relationship that
exists in English-Spanish translations between the translation of fruits and
the translation of their corresponding trees \acite{soleretal93}:
\begin{quote}
\footnotesize
\begin{center}
\begin{tabular}{|l|l| l |l|l|} \cline{1-2} \cline{4-5}
 \multicolumn{2}{|c|}{\bf Fruit} & &  \multicolumn{2}{c|}{\bf Tree}\\ \cline{1-2} \cline{4-5}
{\em English} & {\em Spanish} & & {\em English} & {\em Spanish}\\ \cline{1-2} \cline{4-5} 
almond        & almendra      & & almond tree   & almendro \\ \cline{1-2} \cline{4-5}
apple         & manzana       & & apple tree    & manzano \\ \cline{1-2} \cline{4-5}
cherry        & cereza        & & cherry tree   & cerezo  \\ \cline{1-2} \cline{4-5}
orange        & naranja       & & orange tree   & naranjo \\ \cline{1-2} \cline{4-5}
plum          & ciruela       & & plum tree     & ciruelo \\ \cline{1-2} \cline{4-5}
lemon         & lim\'on       & & lemon tree    & limonero \\ \cline{1-2} \cline{4-5}
\end{tabular}
\end{center}
\end{quote}

The relevant relationship may be described by the following bi-lexical
rule: \\

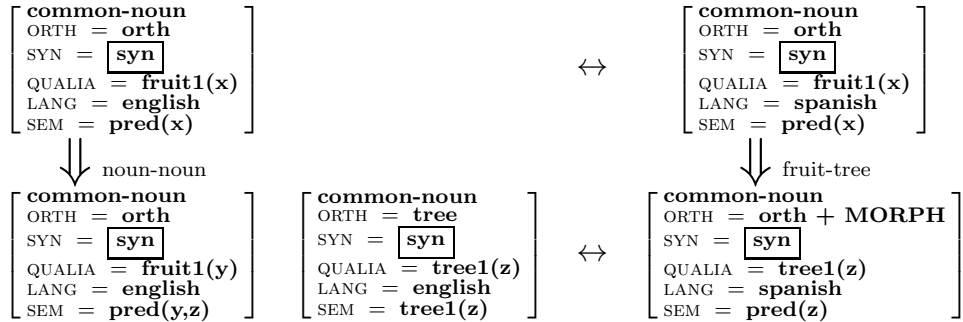
\begin{tabular}{cccc}
{\scriptsize
$\avmplus{\att{common-noun}\\
\attvaltyp{orth}{orth}\\
\attvalshrunktyp{syn}{syn}\\
\attvaltyp{qualia}{fruit1(x)}\\
\attvaltyp{lang}{english}\\
\attvaltyp{sem}{pred(x)}}$} & &
$\leftrightarrow$ &
{\scriptsize $\avmplus{\att{common-noun}\\
\attvaltyp{orth}{orth}\\
\attvalshrunktyp{syn}{syn}\\
\attvaltyp{qualia}{fruit1(x)}\\
\attvaltyp{lang}{spanish}\\
\attvaltyp{sem}{pred(x)}}$} \\
{\LARGE $\Downarrow$} {\scriptsize noun-noun} &  & & {\LARGE $\Downarrow$} {\scriptsize fruit-tree}\\
{\scriptsize
$\avmplus{\att{common-noun}\\
\attvaltyp{orth}{orth}\\
\attvalshrunktyp{syn}{syn}\\
\attvaltyp{qualia}{fruit1(y)}\\
\attvaltyp{lang}{english}\\
\attvaltyp{sem}{pred(y,z)}}$} 
& 
{\scriptsize
$\avmplus{\att{common-noun}\\
\attvaltyp{orth}{tree}\\
\attvalshrunktyp{syn}{syn}\\
\attvaltyp{qualia}{tree1(z)}\\
\attvaltyp{lang}{english}\\
\attvaltyp{sem}{tree1(z)}}$} 
&
$\leftrightarrow$ &
{\scriptsize $\avmplus{\att{common-noun}\\
\attvaltyp{orth}{orth + MORPH}\\
\attvalshrunktyp{syn}{syn}\\
\attvaltyp{qualia}{tree1(z)}\\
\attvaltyp{lang}{spanish}\\
\attvaltyp{sem}{pred(z)}}$}
\end{tabular}

\bigskip
This bi-lexical rule says that if there is a bilexical entry translating
English fruit nouns into Spanish fruit nouns, then there is a bilexical entry
translating `{\em noun} tree' in English into a morphologically derived
tree-denoting noun in Spanish. 

We adopt Qualia structure \acite{pustejovsky91} as our lexical-semantic
representation formalism. According to Pustejovsky, Qualia structure is one of
the four main types of information to be associated with a lexical entry (the
others being Argument, Event and Inheritance structure). The information
incorporated in a Qualia structure specifies the semantics of a lexical item by
virtue of the relations and properties in which it participates. For this
example we assume a simplified Qualia value \acite{pustejovsky91} indicating
whether a noun denotes a tree or a fruit. Note that the morphology of the
output Spanish lexical sign is left implicit since it depends on the actual
noun used (see fruit-tree table above); in addition, the English rule mapping a
noun into a noun modifier is a practical simplification of the complex issue of
noun-noun modification which we do not address here
\acite{pustejovskyetal93,johnstonetal95}. Another
point to note is that we will be vague regarding the amount of information
shared between the input and output lexical signs of lexical rules; a full
treatment of this issue involves aspects of default unification which are
beyond the scope of this paper \acite{meurers94,lascaridesetal95}. Suffice it
to say that in our implementation, an attempt has been made to share maximum
information between input and output lexical signs, although values such
as semantic variables are not shared between input and output lexical signs.

In the abbreviated notation introduced earlier, the above bi-lexical rule will
be represented as:
\begin{quote}
\begin{tabular}{llcl}
\footnotesize
Ne$_{x}$     &             & $\leftrightarrow$ & Ns$_{x}$ \\
$\Downarrow$ {\scriptsize identity} & & & $\Downarrow$ {\scriptsize fruit-tree}\\
Ne$_{y,z}$     & tree1$_{z}$ & $\leftrightarrow$ & Ns$'_{z}$
\end{tabular}
\end{quote}
Given the translation `apple -- {\em
manzana}', for example, the rule would operate as indicated below:
\begin{quote}
\footnotesize
\begin{tabular}{llcl}
apple1$_{x}$  &            & $\leftrightarrow$ & manzana1$_{x}$ \\
$\Downarrow$ {\scriptsize identity} &            & & $\Downarrow$ {\scriptsize fruit-tree}\\
apple1$_{y,z}$ & tree1$_{z}$  & $\leftrightarrow$ & manzano1$_{z}$
\end{tabular}
\end{quote}
Its output is the additional translation relation `apple tree - {\em manzano}'. Similar
translations are achieved for other fruits. 

Clearly this rule should only apply to fruits which grow on trees and not to
fruits such as strawberries which are found on low growing plants. Such
restrictions need to be incorporated in the monolingual lexical signs and
rules. 

Implementationally, bilexical rules may be applied off-line in order to expand
the bilexicon before processing, or they may be applied during transfer 
to extend the bilexicon just sufficiently to enable transfer. We have
opted for the latter approach.

\subsection{Support Verbs}

We now show how bi-lexical rules can be used in the translation of `thirsty',
basing our analysis on the classification of support verbs proposed by
\cite{danlosetal92} for English-French translation. Their proposal, implemented
as part of a Eurotra project, involves transfer at the Interface Structure.
The essence of their approach is similar to that for multi-lexeme translations
given in Section \ref{lex-neu-sec}: the support verb is deleted from the SL
transfer structure, the adjective `thirsty' is translated into the TL noun
({\em sed} in our case), and an appropriate TL support verb is incorporated
into the TL sentence during generation. Information regarding which support
verb a noun requires is encoded in its lexical entry. 

Support verbs can be of five types: neutral (e.g. `is thirsty'), durative (e.g.
`remain thirsty'), inchoative (e.g. `get thirsty'), terminative (e.g. `stop
being thirsty') and iterative (e.g. `be thirty again').  We will consider
neutral support verbs only although the other categories could also be handled
through bi-lexical rules. One difference between the present approach and that
of Danlos et~al.\ is that we equate the noun `thirst' with the noun {\em sed}
in the bilexicon, rather than equating an adjective and a noun, thus factoring
category and support verb differences:
\begin{exquote}
thirst1$_{x}$ $\leftrightarrow$ sed1$_{x}$
\end{exquote}
We believe this reflects more truly the translation relation that exists
between the two lexical items. An English-Spanish bi-lexical rule is then
introduced to derive the adjective on the English side and to include the
neutral support verb `be'; on the Spanish side the support verb {\em tener}, for
the noun {\em sed}, is introduced:
\begin{quote}
\footnotesize
\begin{tabular}{llcll}
 & N$_{x}$                     & $\leftrightarrow$ &  & N[ntrl=tener]$_{x}$ \\
 & $\Downarrow$ {\scriptsize adjective} &      & & $\Downarrow$ {\scriptsize identity} \\
be1$_{e,s,y}$ & A[ntrl=be]$_{y}$ & $\leftrightarrow$ & 
                                     tener1$_{e,s,y}$ & N[ntrl=tener]$_{y}$
\end{tabular}
\end{quote}
Note that we underspecify the support verb for the input English noun to allow `John
has an unquenchable thirst' and similar examples.  The neutral (ntrl) control
verb required by the English adjective is included in its lexical entry's
Qualia structure. Thus, a fuller TFS for `thirsty' is:
\begin{quote}
{\scriptsize
$\avmplus{\att{adjective}\\
\attvaltyp{orth}{thirsty}\\
\attvalshrunktyp{syn}{syn}\\
\attval{qualia}{\avmplus{\att{qualia}\\
       \attval{supp-verbs}{\avmplus{\att{supp-verbs}\\
			   \attvaltyp{ntrl}{be(e,s,y)}\\
			   \attvaltyp{inch}{get(e,s,y)}}}}}\\
\attvaltyp{lang}{english}\\
\attvaltyp{sem}{thirsty1(y)}}$} 
\end{quote}
In designing an appropriate Qualia structure we have added to the roles
proposed by \cite{pustejovsky91} (Constitutive, Formal, Telic and Agentive) in
order to incorporate information necessary for capturing particular phenomena
\acite{johnstonetal95}.

When translating `John is thirsty', the analyser constructs the 
transfer representation:
\begin{quote}
\footnotesize
john1$_{1}$  be1$_{2,1,3}$  thirsty1$_{3}$
\end{quote}
We include the support verb `be' in our representation, even though it
has empty semantics, in order to encode scoping information -- i.e. to 
prevent `John is a painter' translating as `a painter is John'; this rather
{\em ad hoc} solution could be replaced by a mechanism analogous to the labels
used in Underspecified Discourse Representation Theory \acite{reyle95,franketal95}.

During transfer, the bi-lexical rule above is applied to the bi-lexical entry 
for `thirst' to yield:
\begin{quote}
be1$_{e,s,x}$, thirsty1$_{x}$ $\leftrightarrow$ tener1$_{e,s,x}$, sed1$_{x}$
\end{quote}
This multi-lexeme relation is used to translate `is thirsty' into {\em tiene
sed}; a separate entry translates `John' into {\em Juan}.  Bag generation then
ensures that the TL bag yields a sentence which satisfies the constraints
specified by the TL grammar.

The intuitive description of the above process is that we consider `is thirsty'
not to be translatable compositionally, but instead to require a multi-lexeme
translation.  The purpose of bi-lexical rules then is to minimize the
repetition of information in the bi-lexicon while allowing the exploitation of
monolingual lexical processes.

\subsection{Lexicalization Patterns}
\label{lex-pat-sec}

There are other translation phenomena which can be described through the use of
bi-lexical rules. Consider lexicalization patterns for example \acite{talmy85}:
\begin{exquote}
John {\bf swims across} the river.\\
Juan {\bf cruza} el r\'\i o {\bf nadando}.
\end{exquote}
In the English sentence, the main verb encodes manner (i.e. swimming) and motion,
while in Spanish it encodes path (i.e. across) and motion; the remaining
meaning component in each case is expressed through a modifier. Talmy
attributes these distinctions to differences in lexicalization patterns between
the two languages.

A previous approach to such translations has been to introduce the bilexical entries
`swim -- {\em nadar + ando}' and `across -- {\em cruzar}' \acite{beaven92b}.
This approach, however, only implicitly acknowledges that theses two
translations are only appropriate in conjunction, and that separately they
are in fact unintuitive. This not only increases the non-determinism of
transfer and generation, but can increase the likelihood of incorrect
translations for unseen sentences. In the bi-lexical rule view, one relates
verb translations to translations incorporating lexicalization patterns as follows:
\begin{quote}
\footnotesize
\begin{tabular}{llcll}
 V$_{e,s}$  &                & $\leftrightarrow$ &       & V$'_{e,s}$ \\
 $\Downarrow$ {\scriptsize identity} &           &       &       & $\Downarrow$ {\scriptsize gerund} \\
 V$_{f,t}$ & across1$_{f,x}$ & $\leftrightarrow$ & cruzar$_{f,t,x}$ & V[vform= ing]$'_{f,t}$
\end{tabular}
\end{quote}
This rule derives, for every (movement) verb translation, a multi-lexeme
translation which includes `across' as a modifier (we leave the restriction on
verbs to movement events implicit; also, a simplified description of `across'
is assumed \acite{trujillo95}).

Application of this rule to `swim -- {\em nadar}' may be depicted as follows:
\begin{quote}
\footnotesize
\begin{tabular}{llcll}
 swim1$_{e,s}$  &                & $\leftrightarrow$ &       & nadar1$_{e,s}$ \\
 $\Downarrow$ {\scriptsize identity} & &             &       & $\Downarrow$ {\scriptsize gerund} \\
 swim1$_{f,t}$ & across1$_{f,x}$ & $\leftrightarrow$ & cruzar$_{f,t,x}$ & nadando1$_{f,t}$
\end{tabular}
\end{quote}
Lexicalist translation of `John swims across the river' can then proceed by 
translating `swims across' with the output of this rule and the remaining elements
of the input via other bilexical entries.

\subsection{Head Switching}

The phenomenon of head switching in translation can be exemplified by the
following pair of sentences:
\begin{exquote}
John {\bf just} arrived.\\
Juan {\bf acaba de} llegar.
\end{exquote}
The problem with such translations is that the syntactic head in the SL
sentence is not the syntactic head in its translation. This is a major obstacle
for syntactic and even some semantic based translation systems because of the
recursive nature of their transfer representations. 

Head switching has been given a number of solutions in a variety of systems
\acite{kaplanetal89,sadleretal91,russelletal91,whitelock92,kaplanetal93}.
In our framework, the solution is expressed by the following rule
\footnote{We ignore the (complex) issue of tense for this type of example of
head switching; we expect that it can be tackled independently of the present approach.}:
\begin{exquote}
\begin{tabular}{llcll}
 & V$_{e,s}$                     & $\leftrightarrow$ &  & V$'_{e,s}$ \\
 & $\Downarrow$ {\scriptsize identity} &             &  & $\Downarrow$ {\scriptsize infinitive} \\
just1$_{f}$ & V$_{f,t}$ & $\leftrightarrow$ & acabar\_de1$_{f,t,f}$ & V$'_{f,t}$
\end{tabular}
\end{exquote}
Application to the bilexical entry `arrive -- {\em llegar}' results in:
\begin{exquote}
just1$_{f}$ , arrive1$_{f,t}$ $\leftrightarrow$ acabar\_de1$_{f,t,f}$ , llegar1$_{f,t}$
\end{exquote}
Lexicalist translation progresses as before. To exemplify the use of bi-lexical
rules in head switching, we consider translation in embedded contexts in more
detail now. To translate between:
\begin{exquote}
Mary thinks John just arrived.\\
Mar\'\i a piensa que Juan acaba de llegar.
\end{exquote}
the parser constructs the following representation (again, ignoring issues of
scope and quantification):
\begin{exquote}
mary1$_{1}$ , think1$_{2,1,4}$ , john1$_{3}$ , just1$_{4}$ , arrive1$_{4,3}$
\end{exquote}
Assuming appropriate transfer of `Mary' and `John', translation of the embedded
clause obtains as follows. `Thinks' is translated by the following entry:
\begin{exquote}
think1$_{e,s,f}$ $\leftrightarrow$ pensar\_que1$_{e,s,f}$ 
\end{exquote}
In addition, the output of the previous bi-lexical rule serves for multi-lexeme transfer of
`just arrive' to give the incomplete bag:
\begin{exquote}
\{ pensar\_que1$_{2,1,4}$ , acabar\_de1$_{4,3,4}$ , llegar1$_{4,3}$ \}
\end{exquote}
The final result of transfer is the TL bag:
\begin{exquote}
\{ mar\'\i a1$_{1}$ , pensar\_que1$_{2,1,4}$ , juan1$_{3}$ , acabar\_de1$_{4,3,4}$ , 
llegar1$_{4,3}$ \}
\end{exquote}
During generation, {\em acabar de} is made the syntactic head of the sentence
through grammatical constraints in the Spanish grammar. Illustrative rules
might be:
\begin{exquote}
S$_{e,s}$ $\Rightarrow$ NP$_{s}$ VP$_{e,s}$\\
VP$_{e,s,c}$ $\Rightarrow$ Vvp$_{e,s,c}$ VP$_{c}$\\
VP$_{e,s,c}$ $\Rightarrow$ Vs$_{e,s,c}$ S$_{c}$
\end{exquote}
If {\em pensar\_que} has category Vs, and {\em acabar\_de} has category
Vvp, there is only one ordering of the TL bag by which the constraints
indicated by this small grammar can be satisfied, namely, the order given by 
its translation:
\begin{exquote}
Mar\'\i a piensa\_que Juan acaba\_de llegar.
\end{exquote}

It may be noticed that head selection by the TL grammar is possible because the
event semantic constants in {\em acabar\_de} and {\em llegar} are the same. The
consequence of this is that modifiers which apply to `just arrived' and
`arrived' separately will be indistinguishable during TL generation. Avoiding
this problem entails transferring scoping domains for modifiers in order to
constraint generation. However, we have no readily implementable mechanism for
achieving this in LMT as yet.

This concludes our overview of the different translation mismatches that may be
handled through bi-lexical rules. We now consider some unresolved issues
arising from their use.

\section{Bi-lexical Rule Interaction}
\label{bil-com-sec}

One difficulty we have found with  bilexical rules has been their composition.
For example, consider the following translation:
\begin{exquote}
1) John {\bf marched} the soldiers {\bf across} the valley.\\
1$'$) Juan le {\bf hizo cruzar} el valle a los soldados {\bf marchando}.
\end{exquote}
In our framework, two bi-lexical rules should be applied in such cases: one to
construct causative translations \acite{comrie85,levinetal95}:
\begin{exquote}
\begin{tabular}{llcll}
& march1            &  $\leftrightarrow$ &    & marchar1 \\
& $\Downarrow$ {\scriptsize causative} &   &  & $\Downarrow$ {\scriptsize infinitive} \\
& march1$_{causative}$ & $\leftrightarrow$ & hacer1    & marchar1$_{infinitive}$
\end{tabular}
\end{exquote}
The other to deal with differences in lexicalization patterns such as `march
across -- {\em cruzar marchando}'. The problem is that in isolation neither of
these rules could perform the above translation.  Ideally one should be able to
use the output of one as input to the other to derive `march across -- {\em
hacer cruzar marchando}', but this is not possible because both bi-lexical
rules expect a mono-lexeme bilexical entry.

One possible solution is to manually add further bi-lexical rules which incorporate the
composition of other rules:
\begin{exquote}
\begin{tabular}{llcll}
 V              &      & $\leftrightarrow$ &    & V$'$ \\
 $\Downarrow$ {\scriptsize causative} & &   &    & $\Downarrow$ {\scriptsize gerund} \\
 V & across1 & $\leftrightarrow$ & hacer1 cruzar1 & V$'$
\end{tabular}
\end{exquote}
However, this solution leads to a combinatorial explosion in the number of
bi-lexical rules.

The line of work we are investigating combines bi-lexical rules with the
context variables given in Section \ref{lex-var-sec}. There remain problems in
our implementation, however, which will be evident from the following
description.  In our proposed approach either the causative or the
lexicalization pattern bi-lexical rule, or both, incorporate a context variable
in their output bilexical entry.  For example, assume that the variable is
included in the causative rule:
\begin{exquote}
\begin{tabular}{llcll}
 V   &                 & $\leftrightarrow$ &    & V$'$ \\
 $\Downarrow$ {\scriptsize causative} & &  &    & $\Downarrow$ {\scriptsize infinitive} \\
 (V) & & $\leftrightarrow$ & hacer1 & (V$'$)
\end{tabular}
\end{exquote}
This rule says that whenever there is a verb bilexical entry, there is also
an entry which in the context of a causative verb introduces {\em hacer} in the
TL bag. Applying the rule to `march -- {\em marchar}' gives:
\begin{exquote}
\begin{tabular}{lllcll}
& march1                     & $\leftrightarrow$ &  & marchar1 \\
& $\Downarrow$ {\scriptsize causative} &         &  & $\Downarrow$ {\scriptsize infinitive} \\
2) & (march1$_{causative}$) & $\leftrightarrow$ & hacer1 & (marchar1$_{infinitive}$)
\end{tabular}
\end{exquote}
Lexicalist transfer of `march$_{causative}$ ... across ...' via the output of
this rule and that for lexicalization patterns proceeds as follow: the
causative reading of `march' unifies with the context lexical sign in 2) but
is not translated by it. The TL side therefore only contributes {\em hacer}
to the final TL bag. Via the bi-lexical rule given in Section
\ref{lex-pat-sec}, `march across' is transferred such that {\em cruzar} and {\em
marchando} form part of the final TL bag. The result is therefore {\em hacer
cruzar marchando}, which, in combination with the translation of the rest
of the sentence can form the basis for bag generation.

Our main problem is that of resolving conflicts between the syntactic
constraints imposed by each bi-lexical rule. The causative rule requires the
Spanish side to include an infinitive verb, while the lexicalization pattern
rule requires a gerundive verb. Clearly both constraints cannot be satisfied
for the same lexical sign {\em marchar1}. The problem reflects itself in our
proposal in that the rule which includes the contextual pattern must be chosen
carefully. If the lexicalization pattern rule rather than the causative rule
had included the contextual verb lexical sign, the gerundive {\em marchando}
could not have been generated. Instead, a sentence analogous to `John made the
soldiers march crossing the valley' would result, which is perhaps not
desirable. In other words, the conflict between gerundive and infinitive
morphology for `march' is decided manually in advance.  The interaction of such
decisions with other bi-lexical rules therefore might be unpredictable, and
hence is left for further investigation.

\section{Implementation}

The implemented prototype system contains approximately 250 bilexical entries;
this figure includes 20 proper names, 20 multi-lexeme translations and 6
contextual rules. The following translations were done on a SUN Sparc
workstation using Allegro Common Lisp. The time taken to find all possible TL
sentences is given in seconds; total times are for CPU + typical garbage
collection times.
\begin{exquote}
\begin{tabular}{lr} \hline
Translation  & Total (CPU) \\ \hline
John thinks Mary just arrived           \\
Juan piensa\_que Mar\'\i a acaba\_de llegar & 50 (28) \\ \hline
John swam across the river\\
Juan cruz\'o el r\'\i o nadando &  19 (16)  \\ \hline
John marched the soldiers \\
Juan hizo marchar a los soldados    &  19 (17)     \\ \hline
\end{tabular}
\end{exquote}
These timings are only intended to give some idea of the type and stage of our
implementation, rather than reflect the performance of an optimized system.

\section{Conclusion}

We have introduced the mechanism of bi-lexical rules for incorporating lexical
rules in MT. These rules establish correspondences between bilexical entries
such that given the presence of one entry, the existence of another bilexical
entry can be inferred. We presented various phenomena that can be described
using such rules: noun sense extensions, support verbs, lexicalization patterns
and head switching. The rules provide a useful and motivated extension to the
LMT paradigm by providing it with a uniform approach to the description of a number
of translation phenomena.

The problems arising from conflicting constraints imposed by different translation
relations are described, and a partial solution to these was offered involving
the combined use of bi-lexical rules and contextual variables.

Future work could consider implementing Mel'\v cuk's lexical functions
\acite{heylenetal94} in a manner similar to the way bi-lexical rules were used
in the translation of support verbs.

%Refs. \cite{shopen85} \cite{alshawi92} \cite{briscoeetal93a} \cite{farkasetal78} \cite{hinrichsetal94}

\section*{Acknowledgements}

Thanks to two anonymous reviewers for their valuable comments.  The LKB was
implemented by Ann Copestake as part of the ESPRIT ACQUILEX project. Remaining
errors are mine.

\footnotesize
%\bibliographystyle{/home/sol/staff/iat/tex-files/colon}
%\bibliography{ref}

\end{document}